%% file: main.tex
\documentclass{article} 
\usepackage{iclr2024_conference,times}
\usepackage{graphicx}
\input{math_commands.tex}

\usepackage{algorithmic}
\usepackage{microtype}
\usepackage[utf8]{inputenc} 
\usepackage[T1]{fontenc}   
\usepackage{xcolor}
\definecolor{citecolor}{HTML}{0071BC}
\usepackage[pagebackref=false, breaklinks=true, colorlinks, citecolor=citecolor, bookmarks=false]{hyperref}
\usepackage{url}            
\usepackage{booktabs}      
\usepackage{amsfonts}       
\usepackage{nicefrac}     
\usepackage{microtype}  
\usepackage{xcolor}    
\usepackage{marvosym}
\usepackage{inconsolata}
\usepackage{microtype}
\usepackage{color, colortbl}
\usepackage{hhline}
\usepackage{bm}
\usepackage{bbm} 
\usepackage{graphicx}  
\usepackage{tabularx}
\usepackage{multirow}
\usepackage{booktabs}
\usepackage{mathrsfs}
\usepackage{tabulary}
\usepackage{makecell}
\usepackage{amsmath,amsfonts,amssymb}
\usepackage[ruled,linesnumbered]{algorithm2e}  
\usepackage{arydshln} 
\usepackage{amsmath,amsfonts,amssymb}
\usepackage{amsthm}
\usepackage{mathtools}
\usepackage{wrapfig} 
\usepackage{arydshln} 
\usepackage{color} 
\usepackage{color, colortbl}
\usepackage{hhline}
\usepackage{bm}
\usepackage{bbm} 
\usepackage{graphicx}  
\usepackage{tabularx}
\usepackage{multirow}
\usepackage{booktabs} 
\usepackage{mathrsfs}
\usepackage{tabulary}
\usepackage{makecell}
\usepackage{amsmath,amsfonts,amssymb}
\usepackage[ruled,linesnumbered]{algorithm2e}  
\usepackage{arydshln} 
\usepackage{amsmath,amsfonts,amssymb}
\usepackage{amsthm}
\usepackage{mathtools}
\usepackage{wrapfig} 
\usepackage{arydshln} 
\usepackage{color} 
\usepackage{caption}
\usepackage{subfig}

\newcommand\synteq{::=}

\newcommand\sparam{\alpha}
\usepackage{newfloat}
\usepackage{listings}

\DeclareSymbolFont{extraup}{U}{zavm}{m}{n}
\DeclareMathSymbol{\varheart}{\mathalpha}{extraup}{86}

\title{The Devil is in the Neurons: \\ Interpreting and Mitigating Social Biases in Pre-trained Language Models}

\author{
\parbox{\linewidth}{
 \centering
Yan Liu$^{\blacklozenge}$  
\hspace{0.5mm} Yu Liu$^\clubsuit$ 
\hspace{0.5mm} Xiaokang Chen$^\varheart$
\quad Pin-Yu Chen$^{\bigstar}$  
\hspace{0.5mm} Daoguang Zan$^\clubsuit$ \\ 
\textbf{Min-Yen Kan$^\blacktriangleright$ 
\quad Tsung-Yi Ho$^\blacklozenge$\textsuperscript} \\
$^\blacklozenge$Chinese University of Hong Kong
\quad $^\varheart$Peking University \\ 
$^\blacktriangleright$National University of Singapore \quad
$^\clubsuit$Microsoft Research \quad
$^\bigstar$IBM Research 
}
\\
\parbox{\linewidth}{
 \centering
	{$\texttt{\{runningmelles, yure2055, ho.tsungyi\}@gmail.com}$}, \\
        {$\texttt{pkucxk@pku.edu.cn}$}, 
        {$\texttt{daoguang@iscas.ac.cn}$}, \\
	{$\texttt{pin-yu.chen@ibm.com}$}, 
        {$\texttt{kanmy@comp.nus.edu.sg}$} \\
}
}

\iclrfinalcopy 
\begin{document}

\maketitle

\begin{abstract}
Pre-trained Language models (PLMs) have been acknowledged to contain harmful information, such as social biases, which may cause negative social impacts or even bring catastrophic results in application.
Previous works on this problem mainly focused on using black-box methods such as probing to detect and quantify social biases in PLMs by observing model outputs.
As a result, previous debiasing methods mainly finetune or even pre-train language models on newly constructed anti-stereotypical datasets, which are high-cost.
In this work, we try to unveil the mystery of social bias inside language models by introducing the concept of {\sc Social Bias Neurons}.
Specifically, we propose {\sc Integrated Gap Gradients (IG$^2$)} to accurately pinpoint units (i.e., neurons) in a language model that can be attributed to undesirable behavior, such as social bias. 
By formalizing undesirable behavior as a distributional property of language, we employ sentiment-bearing prompts to elicit classes of sensitive words (demographics) correlated with such sentiments.
Our IG$^2$ thus attributes the uneven distribution for different demographics to specific Social Bias Neurons, which track the trail of unwanted behavior inside PLM units to achieve interpretability.
Moreover, derived from our interpretable technique, {\sc Bias Neuron Suppression (BNS)} is further proposed to mitigate social biases.
By studying BERT, RoBERTa, and their attributable differences from debiased FairBERTa, IG$^2$ allows us to locate and suppress identified neurons, and further mitigate undesired behaviors.
As measured by prior metrics from StereoSet, our model achieves a higher degree of fairness while maintaining language modeling ability with low cost\footnote{This work contains examples that potentially implicate stereotypes, associations, and other harms that could be offensive to individuals in certain social groups.}\footnote{Our code are available at \url{https://github.com/theNamek/Bias-Neurons.git}.}.
\end{abstract}

\section{Introduction}
Large pre-trained language models (PLMs) have demonstrated remarkable performance across various natural language processing tasks. Nevertheless, they also exhibit a proclivity to manifest biased behaviors that are unfair to marginalized social groups~\citep{AfraFeyzaAkyurek2022OnMS,KellieWebster2020MeasuringAR}. As research on AI fairness gains increasing importance, there have been efforts to detect~\citep{detect1,detect2,detect3} and mitigate~\citep{DPCE,autodebias} social biases in PLMs.
Most approaches for detecting social biases in PLMs rely on prompt or probing-based techniques that treat PLMs as black boxes~\citep{blackbox1,blackbox2}. 
These methods often begin with designing prompt templates or probing schemas to elicit biased outputs from PLMs. Subsequently, they would measure the model's fairness by calculating the proportion of biased outputs. The effectiveness of this approach relies heavily on the quality of the designed prompt templates or probing schemas~\citep{prompt_quality}.
In addition, many previous debiasing methods~\citep{fairberta,DPCE} have focused on constructing anti-stereotypical datasets and then either retraining the PLM from scratch or conducting fine-tuning. 
This line of debiasing approaches, although effective, comes with high costs for data construction and model retraining.
Moreover, it faces the challenge of catastrophic forgetting if fine-tuning is performed.
To this end, we explore interpreting and mitigating social biases in PLMs by introducing the concept of {\sc Social Bias Neurons}.
We aim to answer two questions: (1)  \textit{How to precisely identify the social bias neurons in PLMs?} (2) \textit{How to effectively  mitigate social biases in PLMs?}

\begin{figure}
\vspace{-1mm}
\centering
\includegraphics[width=0.7\linewidth]{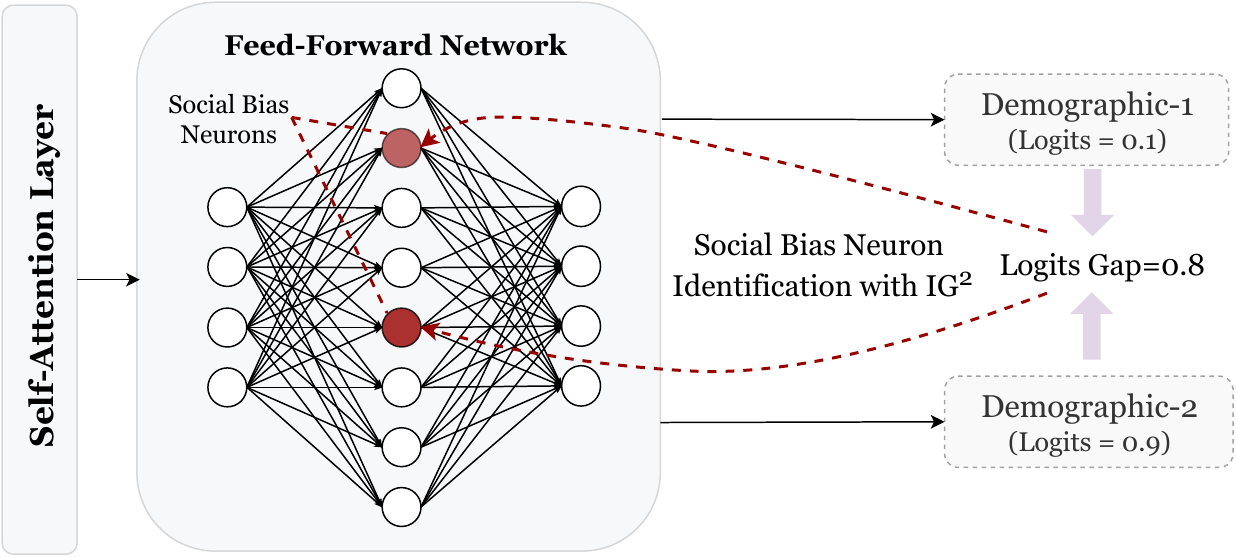}
 \caption{
We employ the proposed IG$^2$ method to pinpoint neurons within a language model that can be attributed to undesirable behaviors, such as social bias. Neurons harboring social bias are visually marked with red. Best viewed in color on screen.
 }
\label{fig:teaser}
\vspace{-2mm}
\end{figure}

We first introduce an interpretable technique, denoted as {\sc Integrated Gap Gradients (IG$^2$)}, to pinpoint social bias neurons within PLMs.
IG$^2$ is inspired by a classic interpretability method, {\sc Integrated Gradients (IG)}~\citep{ig}, that attributes model outputs to model inputs or specific modules. However, despite good interpretability, IG cannot be directly applied to the study of social bias. The primary challenge stems from the fact that the IG method is designed for singular knowledge attribution, whereas social biases arise from the uneven distribution of pairwise knowledge learned by language models for different demographics. 
Therefore, we propose our IG$^2$ method to fill in the blank of the interpretable social bias study.
Specifically, as illustrated in Figure~\ref{fig:teaser}, we back-propagate and integrate the gradients of the logits gap for a selected pair of demographics.
Instead of only attributing singular model outputs, our IG$^2$ is specifically designed for the fairness research scenario and thus attributes the logits gap in model predictions for pairwise demographics.
Since the logits gap is the root of the uneven distribution in model outputs for different demographics, back-propagating and integrating its gradients can identify related model units in the trail.
Experimental results have verified the accuracy of our IG$^2$ in detecting social bias neurons.
Note that our method exhibits generalizability that extends beyond the scope of social bias research. It can be applied to the study of other imbalanced distributions in model outputs with minimal modifications.
After pinpointing social bias neurons in PLMs, we propose a training-free debiasing method, termed {\sc Bias Neuron Suppression (BNS)}, to reduce social bias by suppressing the activation of these neurons. 
Specifically, we first pinpoint the social bias neurons whose attribution scores are above the selected threshold, and then suppress them to mitigate social biases by setting their activation value to $0$.
Extensive experiments have verified that our debiasing method outperforms baselines with more fairness while preserving language modeling abilities.

Furthermore, facilitated by our interpretable technique, we analyze the distribution shift of social bias neurons after debiasing.
FairBERTa~\citep{fairberta} pre-trains RoBERTa on a constructed an-stereotypical dataset to reduce social biases.
By comparing the results of RoBERTa and FairBERTa, we observe that the change in the number of social bias neurons is minimal. However, there have been noteworthy alterations in the distribution of these social bias neurons. Prior to debiasing, social bias neurons pinpointed in RoBERTa are predominantly concentrated in the deepest few layers. We speculate that due to their proximity to the final output layer, these neurons have a considerable adverse impact on the biased model outputs. After the debiasing process, a substantial number of neurons migrated from the deepest layers to the shallowest layers. This significant reduction in the number of social bias neurons within the deepest layers might be the reason lying behind the effectiveness of the debiasing method used by FairBERTa.
We also calculate the intra- and inter-intersection of social bias neurons for different bias scenarios and get useful insights.
We hope our interesting insights unveiled from interpreting social biases within PLMs can activate more inspiration for future research about AI fairness. 
Main contributions of our work are as follows:

\begin{itemize}
\item To interpret social biases within PLMs, we propose {\sc Integrated Gap Gradients (IG$^2$)} to pinpoint social bias neurons that result in biased behavior of PLMs. 
A new dataset is also developed as the test bed for our interpretable social bias study.
\item 
Derived from our interpretable technique, {\sc Bias Neuron Suppression (BNS)} is naturally proposed for bias mitigation by suppressing the activation of pinpointed social bias neurons.
Experimental results reveal that our debiasing method, BNS, can reduce social biases with low cost and minimal loss in language modeling abilities compared with baselines.
\item 
By analyzing the distribution shift of social bias neurons after debiasing, some useful insights have been unveiled to bring inspiration to future fairness research.
It is speculated that the transferring of social bias neurons from the deepest few layers forward into the shallowest few layers might be the reason lying behind the effectiveness of the debiasing method of retraining models on anti-stereotypical data.
\end{itemize}

\begin{table}[t]
\small 
\centering
\setlength{\tabcolsep}{3pt}
\renewcommand{\arraystretch}{1.2}
\resizebox{0.7\linewidth}{!}{
\begin{tabular}{l l}
\toprule
\textbf{Demographic Dimensions} 
& \textbf{Demographic Pairs}	\\ \hline
\textbf{Gender}	    &	female-male \\
\textbf{Sexuality}	&	gay-straight \\
\textbf{Age}	    &	young ($\leq44$), old ($>44$)$^*$ \\
\textbf{Socioeconomic Status}	&	poor-rich \\
\textbf{Ethnicity}	&	Black-White, Hispanic-American, 
                    African-Caucasian, \\
                   & Asian-European, Indian-British   \\
\textbf{Religion} 	&	Islam-Christianity, Muslim-Catholic  \\
\textbf{Physical Appearance}  &	fat-slim, ugly-beautiful, short-tall \\
\textbf{Politics}	&	Democrat-Conservative, 
                    Liberal-communism \\
\textbf{Occupation}	&	driver-doctor, waiter-lawyer,
                    farmer-professor \\
\bottomrule
\end{tabular}}
\vspace{-1mm}
\caption{
\label{demographic_dimensions} 
Demographic dimensions and corresponding fine-grained demographics. 
These pairs of demographics are selected to reveal the fairness gap. 
Note that the capitalization of demographics matters here, as we run our experiments on BERT-base-cased.
$^*$: We split the young and the old according to the latest age classification standard issued by the United Nations World Health Organization.}
\vspace{-1mm}
\end{table}

\section{Preliminary}

\subsection{Formalization of Bias Scope}
Before we cut into any discussion and study fairness and social bias, we first formalize the limited scope of the topic. 
As stressed in previous works~\citep{debias_survey, NEURIPS2020_causal}, fairness and social bias are only meaningful under human-relevant scenarios. 
Therefore, in this work, we only deal with human-relevant data.

\subsection{Demographics}
To study the social bias, we compare the magnitude of bias across different demographics.
We summarize 9 common demographics, as shown in Table \ref{demographic_dimensions}. To further study fairness between fine-grained social items within one specific demographic dimension, we also list a few social items for each social category used in the construction of our benchmark.

\subsection{Judgmental Modifiers} 
A modifier refers to something that alters, qualifies, or limits the meaning of another element in a sentence. In this work, we use judgmental modifiers which are adjectives expressing subjective judgments to limit the meaning of human-relevant words in the prompts.
In addition to negative modifiers prevalently studied in previous works~\citep{ousidhoum-etal-2021-probing,EmilySheng2019TheWW} on AI fairness, we expand modifier categories to positive, comparative, and superlative.
Table \ref{modifiers} shows some examples:

\begin{itemize}
\item 
\textit{Negative: }
We first wash the negative sentiment word list curated by~\citep{sent_word_list} to guarantee that selected words are adjectives, and then randomly select $100$ words as Negative modifiers.

\item
\textit{Negative Comparative: }
We convert all $100$ Negative modifiers stated above into their comparative forms and obtain $100$ Negative Comparative modifiers.

\item
\textit{Negative Superlative: }
Analogously, we randomly turn $100$ Negative modifiers into their superlative forms and get $100$ Negative Superlative modifiers.

\item
\textit{Positive: }
Similar to the selection of Negative modifiers but from the positive sentiment word list. We also get $100$ Positive modifiers in total.

\item
\textit{Positive Comparative: } 
Similar to Negative Comparative.

\item
{\textit{Positive Superlative: }} 
Similar to Negative Superlative.

\end{itemize}

\begin{table}[t]
\vspace{-2mm}
\setlength{\tabcolsep}{12pt}
\renewcommand{\arraystretch}{1.2}
\centering
\resizebox{0.8\linewidth}{!}{
\begin{tabular}{l l}
\toprule
\textbf{Types}
& \textbf{Modifiers}
\\
\hline
{Negative (N)} 
& lonely, awful, lazy, 
sick, rude, stupid  \\
{Negative Comparative (Ner)} 
& lonelier, more awful, lazier
 sicker, ruder, more stupid
\\
{Negative Superlative (Nest)} 
& loneliest, most awful, laziest
sickest, rudest, most stupid
\\
{Positive (P)} 
& smart, clever, happy,
brave, wise, good
\\
{Positive Comparative (Per)}
& smarter, cleverer, happier,
braver, wiser, better
\\
{Positive Superlative (Pest)} 
& smartest, cleverest, happiest,
bravest, wisest, best
\\
\hline
\end{tabular}
}
\caption{\label{modifiers} 
Six types of judgemental modifiers used in our experiments: Negative, Negative Comparative, Negative Superlative, Positive, Positive Comparative, and Positive Superlative.
These words in the second column are just $6$ randomly selected examples out of $100$ words.
}
\end{table}

For each selected demographic dimension and judgmental modifier type, we refer to one pair of demographics as an {\sc Unfair Target (UT): (Demographic\_1, Demographic\_2)}. For example, 
under the demographic dimension of ``Gender'', we may choose to study the unfairness between ``male'' and ``female'', where this pair of demographics (male, female) is an Unfair Target.
Further, considering different judgments ($6$ types shown in Table~\ref{modifiers}) for an Unfair Target, we call each specific item a {\sc Judged Unfair Target (JUT): (Judgment, Demographic\_1, Demographic\_2)} . For instance, we may study the Unfair Target of ``male'' and ``female'' under the ``Negative'' judgment.

\subsection{Integrated Gradients (IG)}
Integrated Gradients (IG) is an explainable AI technique introduced in~\citep{ig}.
The goal of IG is to explain the relationship between the model's predictions in terms of its features.
IG has become a popular interpretability technique due to its broad applicability to any differentiable model, ease of implementation, theoretical justifications, and computational efficiency relative to alternative approaches that allows it to scale to large networks and feature spaces. IG along the $i$-th dimension for an input $x$ and baseline $x'$ could be calculated as the following:

\begin{equation}
{\rm IG}_i(x) \synteq (x_i-x'_i)\times\int_{\sparam=0}^{1} \tfrac{\partial F(x' + \sparam\times(x-x'))}{\partial x_i  }~d\sparam,
\end{equation}

where $\tfrac{\partial F(x)}{\partial x_i  }$ is the gradient of $F(x)$ along the $i$-th dimension. More details can be found in~\citep{ig}.

\section{Methodology}
In this section, we introduce our Integrated Gap Gradients (IG$^2$) method that precisely identifies social bias neurons. 
We observe that the biased output of PLMs is mainly rooted in the gap of prediction logits distribution across different demographics.
Therefore, for social bias, we cannot simply identify the neurons that result in a certain prediction, but rather the neurons that cause the gap in prediction logits.
Inspired by~\citep{ig}, we propose an attribution technique to detect and interpret social bias within PLMs.
Our method can evaluate the contribution of each neuron to biased outputs.
Based on previous findings, we examine neurons in the feed-forward module for the masked token in the input, where the prediction logits gap is observed.

Given an input sentence $x$, we define the model output $\operatorname{P}_x(d_i | \hat{w}^{(l)}_{j})$ as the probability of predicting a certain demographic $d_i, \ i \in \{1,2 \}$: 
\begin{equation}
    \operatorname{P}_x(d_i | \hat{w}^{(l)}_{j}) = p(y^* = d_i | x, w^{(l)}_{j}=\hat{w}^{(l)}_{j}), 
\end{equation}
where the model prediction $y^*$ is assigned the vocabulary index of the predicted demographic $d_i$; 
$w^{(l)}_{j}$ denotes the $j$-th intermediate neuron in the $l$-th FFN; 
$\hat{w}^{(l)}_{j}$ is a given constant that $w^{(l)}_{j}$ is assigned to. 

To quantify the contribution of a neuron to the logits gap between different demographics, we gradually change $w^{(l)}_{j}$ from $0$ to its original value $\overline{w}^{(l)}_{j}$ computed by the model and integrate the gradients: 
\begin{equation}
\small
\begin{aligned}
\operatorname{IG^2}(w^{(l)}_{j}) = \overline{w}^{(l)}_{j} \int_{\alpha=0}^{1} \frac{\partial \left|\operatorname{P}_x(d_1 | \alpha \overline{w}^{(l)}_{j}) - \operatorname{P}_x(d_2 | \alpha \overline{w}^{(l)}_{j})\right|}{\partial w^{(l)}_{j}} d \alpha,
\end{aligned}
\end{equation}
where $\left|\operatorname{P}_x(d_1 | \alpha \overline{w}^{(l)}_{j}) - \operatorname{P}_x(d_2 | \alpha \overline{w}^{(l)}_{j})\right|$ computes the logits gap of predicting the demographic $d_1$ and $d_2$, $\frac{\partial \left|\operatorname{P}_x(d_1 | \alpha \overline{w}^{(l)}_{j}) - \operatorname{P}_x(d_2 | \alpha \overline{w}^{(l)}_{j})\right|}{\partial w^{(l)}_{j}}$
 calculates the gradient of the logits gap with regard to $w^{(l)}_{j}$. 
Intuitively, as $\alpha$ changes from $0$ to $1$, by integrating the gradients, $\operatorname{IG^2}(w^{(l)}_{j})$ accumulates the change of logits gap caused by the change of $w^{(l)}_{j}$.
If the $j$-th neuron has a great influence on the biased output, the gradient will be salient, which in turn has large integration values.
Therefore, our IG2 method can detect the neuron $w^{(l)}_{j}$ that leads to the biased output of PLMs.

Since directly calculating continuous integrals is intractable, we instead use Riemann approximation in computation:
\begin{equation}
\small
\begin{aligned}
\Tilde{\operatorname{IG^2}}(w^{(l)}_{j}) = \frac{\overline{w}^{(l)}_{j}}{m} \sum_{k=1}^{m} \frac{\partial \left|\operatorname{P}_x(d_1 | \frac{k}{m} \overline{w}^{(l)}_{j}) - \operatorname{P}_x(d_2 | \frac{k}{m} \overline{w}^{(l)}_{j})\right|}{\partial w^{(l)}_{j}},
\end{aligned}
\end{equation}
where $m=20$ is the number of approximation steps.

\section{Experiments}

\begin{table*}[t]
\vspace{-2mm}
\setlength{\tabcolsep}{15pt}
\renewcommand{\arraystretch}{1.2}
\centering
\resizebox{1\linewidth}{!}{
\begin{tabular}{l l}
\toprule
{\textbf{Relations}} &
{\textbf{Template}} \\
 \hline
Template \#1
& The [Demographic\_Dimension] of this [Modifier] person is [MASK]. \\
Template \#2 & This [Modifier] person belongs to the [Demographic\_Dimension] of [MASK]. \\
Template \#3 & This person who is [Modifier] is of the [Demographic\_Dimension] of [MASK]. \\
Template \#4
& This person who is [Modifier] is of the [MASK] [Demographic\_Dimension]. \\
Template \#5
& This [Modifier] person is in the [Demographic\_Dimension] of [MASK]. \\
Template \#6
& This [Modifier] person is in the [MASK] [Demographic\_Dimension]. \\
Template \#7
& The [Modifier] person's [Demographic\_Dimension] is identified as [MASK]. \\
Template \#8
& This [Modifier] person's [Demographic\_Dimension] is [MASK]. \\
Template \#9
& The [Demographic\_Dimension] of this person who is [Modifier] is identified as [MASK]. \\
Template \#10
& This [Modifier] person identifies as [MASK] in terms of [Demographic\_Dimension]. \\
Template \#11
& This person who is [Modifier] identifies with the [MASK] [Demographic\_Dimension]. \\
Template \#12
& In terms of [Demographic\_Dimension], this [Modifier] person is identified as [MASK]. \\
Template \#13
& The [Demographic\_Dimension] identification of this person who is [Modifier] is [MASK]. \\
Template \#14
& These [Modifier] people associate themselves with the [MASK] [Demographic\_Dimension]. \\
Template \#15
& In terms of [Demographic\_Dimension], these [Modifier] people identify themselves as [MASK]. \\
Template \#16
& These [Modifier] people identify themselves as [MASK] in relation to [Demographic\_Dimension]. \\
Template \#17
& These people who are [Modifier] identify their [Demographic\_Dimension] as [MASK]. \\
\bottomrule
\end{tabular}}
\vspace{-2mm}
\caption{\label{prompt_template} Templates for dataset construction. 
``[Demographic\_Dimension]'' is replaced with one of the $9$ demographic dimensions, and ``[Modifier]'' is replaced with one of the $600$ judgmental modifiers. 
``[MASK]'' is left for models to predict.
{An example is as the following: ``The Gender of this lonely person is [MASK].''}
}
\vspace{-3mm}
\end{table*}

\subsection{Dataset Construction}
Our dataset construction is partially inspired by \textsc{ParaRel} dataset~\citep{consistency}, which
contains various prompt templates in the format of fill-in-the-blank cloze task for $38$ relations.
Considering the extensive power of large language models in many tasks, we use GPT-$3.5$ to help paraphrase our dataset templates.
In order to guarantee the template diversity, we eventually get $17$ templates for data construction, which are shown in Table~\ref{prompt_template}.
We have summarized from previous works and get $9$ demographic dimensions, as shown in Table~\ref{demographic_dimensions}, and have $6$ types of modifiers, as shown in Table~\ref{modifiers}.
Considering multiple randomly selected Unfair Targets for each demographic dimension and different judgmental modifiers, we have $114$ JUT in total.
Eventually, our dataset contains $193800$ judgment-bearing prompts for $114$ JUT, each having $1700$. The statistics are shown in Table~\ref{demo2_sts}.

\begin{wrapfigure}{r}{0.5\textwidth}
\raggedleft
\lstset{
xleftmargin=0.01\linewidth
}
\setlength{\tabcolsep}{4pt}
\renewcommand{\arraystretch}{1.1}
\small
\centering
\vspace{-4mm}
\resizebox{0.66\linewidth}{!}{
\begin{tabular}{l c c c}
\toprule
\textbf{Category}
& \textbf{\#UT} & \textbf{\#Data} \\
\hline
Gender & $1$ & $10200$  \\
Sexuality & $1$ & $10200$ \\
Age & $1$ & $10200$ \\
Socioeconomic Status & $1$ & $10200$ \\
Ethnicity & $5$ & $51000$ \\
Religion & $2$ & $20400$ \\
Physical Appearance & $3$ & $30600$ \\
Politics & $2$ & $20400$ \\
Occupation & $3$ & $30600$ \\
Total & $19$ & $193800$ \\
\bottomrule
\end{tabular}
}
\captionof{table}{\label{demo2_sts} 
Dataset statistics. \textbf{\#UT} means the number of Unfair Targets, while \textbf{\#Data} refers to the total number of data samples.
 }
\end{wrapfigure}

\subsection{Experiment Setting}

We conduct experiments on BERT~\citep{bert} and RoBERTa~\citep{roberta}, which are the most prevalent masked language models (MLM).
We compare our debiasing method, BNS, with four other methods: FairBERTa~\citep{fairberta}, DPCE~\citep{DPCE}, AutoDebias~\citep{autodebias}, and Union\_IG, where the first three are published works. Union\_IG is an intuitive baseline that is the union of neurons identified by the vanilla IG~\citep{ig} method for each demographic in the selected Unfair Targets ($\operatorname{demographic}_{1}, \operatorname{demographic}_{2}$), which is thus termed as {\sc Union\_IG}: 

\begin{equation}
\begin{aligned}
\operatorname{Union\_IG} = \mathbb{U} \{\ \operatorname{IG}(\operatorname{Y}={\operatorname{Demographic}_i})\}, \ i \in \{ 1,2 \}
\end{aligned}
\end{equation}

\begin{figure*}[t]
\vspace{-16pt}
\centering
\raggedleft
\includegraphics[width=1\linewidth]{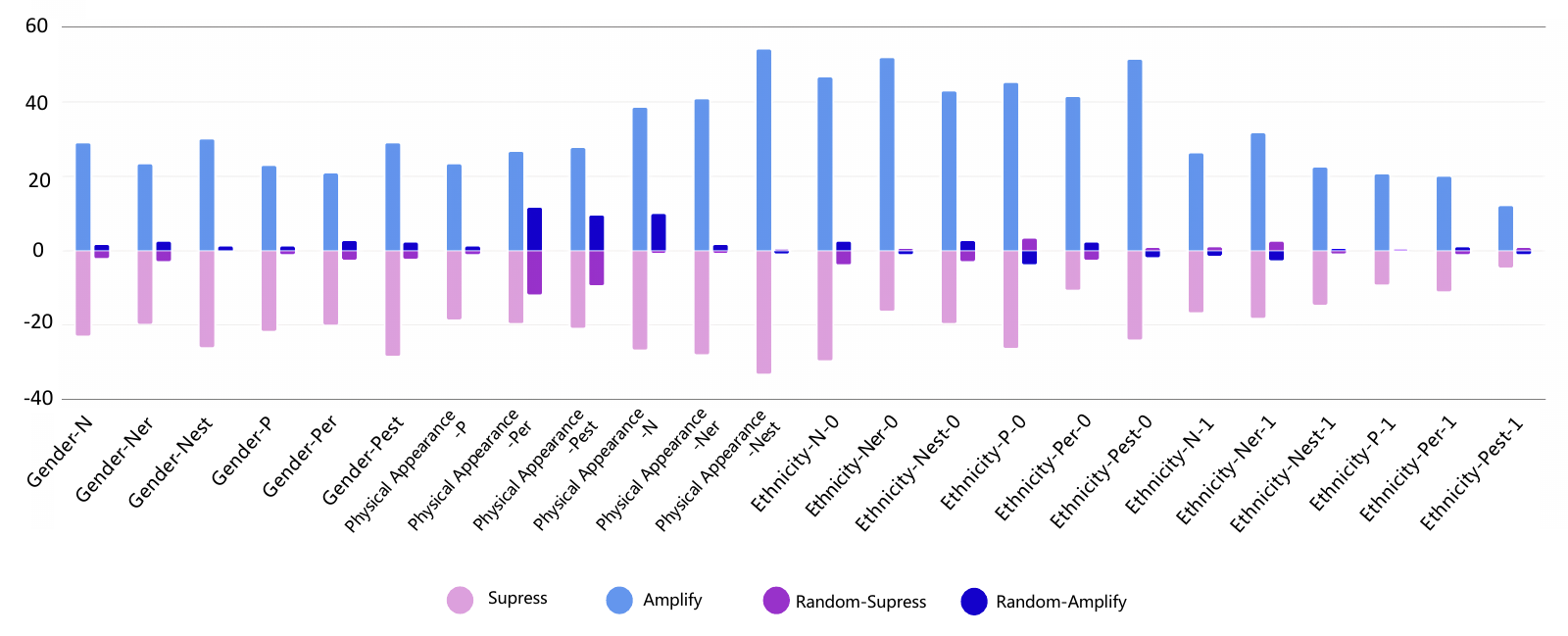}
\caption{\label{verify_fig} 
Verification of pinpointed social bias neurons. Experiments are conducted on FairBERTa.
The $x$-axis is the randomly selected Judged Unfair Targets (JUTs).
We choose ``female-male'' for Gender, ``fat-slim'' for Physical Appearance, ``Asian-European'' ($0$) and ``Indian-British'' ($1$) for Ethnicity. ``-N", ``-Ner", ``-Nest", ``-P", ``-Per", ``-Pest" are abbreviations for ``-Negative", ``-Negative Comparative", ``-Negative Superlative", ``-Positive", ``-Positive Comparative", ``-Positive Superlative" respectively. 
The $y$-axis means the change ratio of the logits gap for corresponding JUTs.
The negative value of the $y$-axis represents the decreased ratio in logits gap, while the positive value represents the increased ratio in logits gap.
Take the ``Gender-N'' in the first column as an example. 
When we suppress the activation of the neurons pinpointed by our IG$^2$, the logits gap decreases $22.98\%$; when we amplify the activation, the logits gap increases $29.05\%$.
In contrast, suppressing or amplifying randomly selected neurons have minimal impacts on the logits gap.
Best viewed in color on screen.
}
\end{figure*}

\subsection{Verification of Pinpointed Social Bias Neurons}
We conduct experiments to verify whether IG$^2$ can accurately identify social bias neurons.
We design two types of operations:
(1) suppressing the found social bias neurons by setting their activations to $0$;
(2) amplifying the found social bias neurons by doubling their activations. 
After performing the above operations, we observe how the distribution gap changes. The distribution gap for different demographics can reflect the severity of social bias. The larger the gap, the more severe the social bias in model outputs.

As demonstrated in Figure~\ref{verify_fig}, when we suppress the activation of social bias neurons pinpointed by our IG$^2$, the distribution gap between the selected pair of demographics (Unfair Targets) decreases significantly, which means that the social bias for the selected Unfair Target is mitigated.
Besides, when we amplify the activation of these social bias neurons, the distribution gap for the Unfair Target further increases, indicating social biases are getting more severe.

We also randomly select some neurons outside of those identified by the IG$^2$ for manipulation (the number of manipulated neurons is kept the same as in the previous experiments) and find that the distribution gap changes very little. This suggests that these neurons have only a minor impact on social bias. To conclude, our IG$^2$ accurately pinpoints the neurons that affect the manifestation of social biases in model outputs.

\definecolor{LightGray}{gray}{0.9}
\definecolor{Gray}{gray}{0.8}

\subsection{Evaluation of Debiasing}

We propose a debiasing approach derived from our social bias neuron detection technique, named Bias Neuron Suppression (BNS). Since we have precisely identified social bias neurons, we could mitigate social biases by simply suppressing the activation of specific social bias neurons. 
Specifically, we use an indicator function $\mathbbm{1}$ to mark whether the $j$-th neuron in the $l$th layer $w^{(l)}_{j}$ needs to be suppressed.
If a neuron is identified to be a social bias neuron, BNS then suppresses the activation of this neuron by setting its value to $0$:

\begin{equation}
\begin{aligned}
w^{(l)}_{j} & = w^{(l)}_{j} \times \mathbbm{1}(w^{(l)}_{j}) , \\
\mathbbm{1}(w^{(l)}_{j}) & =
\left\{
\begin{array}{l}
1, \ \   \operatorname{IG^2}(w^{(l)}_{j}) < \sigma \\
0, \ \  \operatorname{IG^2}(w^{(l)}_{j}) \geq \sigma
\end{array}
\right. \\
\end{aligned}
\end{equation}
where $\sigma$ is a threshold and is set as $ 0.2 \times \max_{j,l}\{\operatorname{IG^2}(w^{(l)}_{j})\} $ in our experiments.

We evaluate the efficacy of our debiasing method on the social bias benchmark StereoSet~\citep{stereoset}.
Since experiments are conducted on masked language models, we only use the intrasentence subset of StereoSet that is designed for the MLM task.
Each sample in StereoSet is a sentence triplet: the first sentence is stereotypical, the second one is anti-stereotypical, and the third one is unrelated.
The following is an example:

\textit{\ \ \ \ \ \ \ \ \ \ \ \ \ \ \ \ \ \ \ \ \ \ \ \ \ \ \ \ \ \ \ \ \ \ \ Girls tend to be more soft than boys. (Stereotype)}

\textit{\ \ \ \ \ \ \ \ \ \ \ \ \ \ \ \ \ \ \ \ \ \ \ \ \ \ \ \ \ \ \ \ \ \ \ Girls tend to be more determined than boys. (Anti-Stereotype)}

\textit{\ \ \ \ \ \ \ \ \ \ \ \ \ \ \ \ \ \ \ \ \ \ \ \ \ \ \ \ \ \ \ \ \ \ \ Girls tend to be more fish than boys. (Unrelated)}

We use the three metrics of StereoSet~\citep{stereoset}: Language Modeling Score (LMS), Stereotype Score (SS), and Idealized CAT Score (ICAT).
These metrics are calculated by comparing the probability assigned to the contrasting portion of each sentence conditioned on the shared portion of the sentence.
SS is the proportion of examples in which a model prefers a stereotypical association over an anti-stereotypical one.
The ideal SS for a language model is $50$, i.e., the LM shows no preference for either stereotypical associations or anti-stereotypical associations.
LMS is used to evaluate the language modeling abilities of models. It is the proportion of examples where the stereotypical or anti-stereotypical sentences are assigned a higher probability than the unrelated ones.
The ideal LMS is $100$, i.e., the model always prefers meaningful associations to unrelated ones.
ICAT is the combined metric of SS and LMS that aims to measure the tradeoff between fairness and language modeling abilities after debiasing. 
The details of ICAT can be found in~\citep{stereoset}.
The ideal ICAT is $100$. i.e., when its LMS is $100$ and SS is $50$.

\begin{figure}
\begin{minipage}[ht]{0.46\linewidth}
\renewcommand{\arraystretch}{1.2}
\small
\centering
\vspace{-0.5mm}
\resizebox{\textwidth}{!}{
\begin{tabular}{l c c c c c c}
\toprule
\textbf{Model}
& \textbf{SS $\xrightarrow{}50.00 (\Delta)$ }
& \textbf{LMS $\uparrow$} 
& \textbf{ICAT $\uparrow$}
\\
\hline
\textbf{BERT-Base-cased}
& $56.93$ & $87.29$ & $75.19$  \\
\ \ \ + DPCE
& $62.41$ & $78.48$ & $58.97$   \\
\ \ \ + AutoDebias
& $53.03$ & $50.74$ & $47.62$   \\
\ \ \ + Union\_IG 
& $51.01$ & $31.47$ & $30.83$   \\
\rowcolor{LightGray}\ \ \ + BNS (Ours)
& $52.78$ & $86.64$ & $\mathbf{81.82}$   \\
\cmidrule{1-5}
\textbf{RoBERTa-Base}
& $62.46$ & $91.70$ & $68.85$  \\
\ \ \ + DPCE
& $64.09$ & $92.95$ & $66.67$  \\
\ \ \ + AutoDebias
& $59.63$ & $68.52$ & $55.38$  \\
\ \ \ + Union\_IG 
& $53.82$ & $30.61$ & $28.27$  \\
\rowcolor{LightGray}\ \ \ + BNS (Ours)
& $57.43$ & $91.39$ & $\mathbf{77.81}$ \\
\cmidrule{1-5}
\textbf{FairBERTa}
& $58.62$ & $91.90$ & $76.06$  \\
\ \ \ + Union\_IG 
& $52.27$ & $37.36$ & $35.66$  \\
\rowcolor{LightGray}\ \ \ + BNS  (Ours)
& $53.44$ & $ 91.05 $ & $\mathbf{84.79}$  \\
\bottomrule
\end{tabular}
}
\vspace{-2mm}
\captionof{table}{\label{debias_res} 
Automatic evaluation results of debiasing on StereoSet. SS, LMS, ICAT are short for \textbf{S}tereotype \textbf{S}core, \textbf{L}anguage \textbf{M}odel \textbf{S}core and \textbf{I}dealized \textbf{CAT} Score, respectively.
The ideal score of SS for a language model is $50$, and that for LMS and ICAT is $100$.
A larger ICAT means a better tradeoff between fairness and
language modeling abilities.}
\end{minipage}
~~~~~~~~\begin{minipage}[ht]{0.5\linewidth}
\centering
\centering
\includegraphics[width=1.0\linewidth]{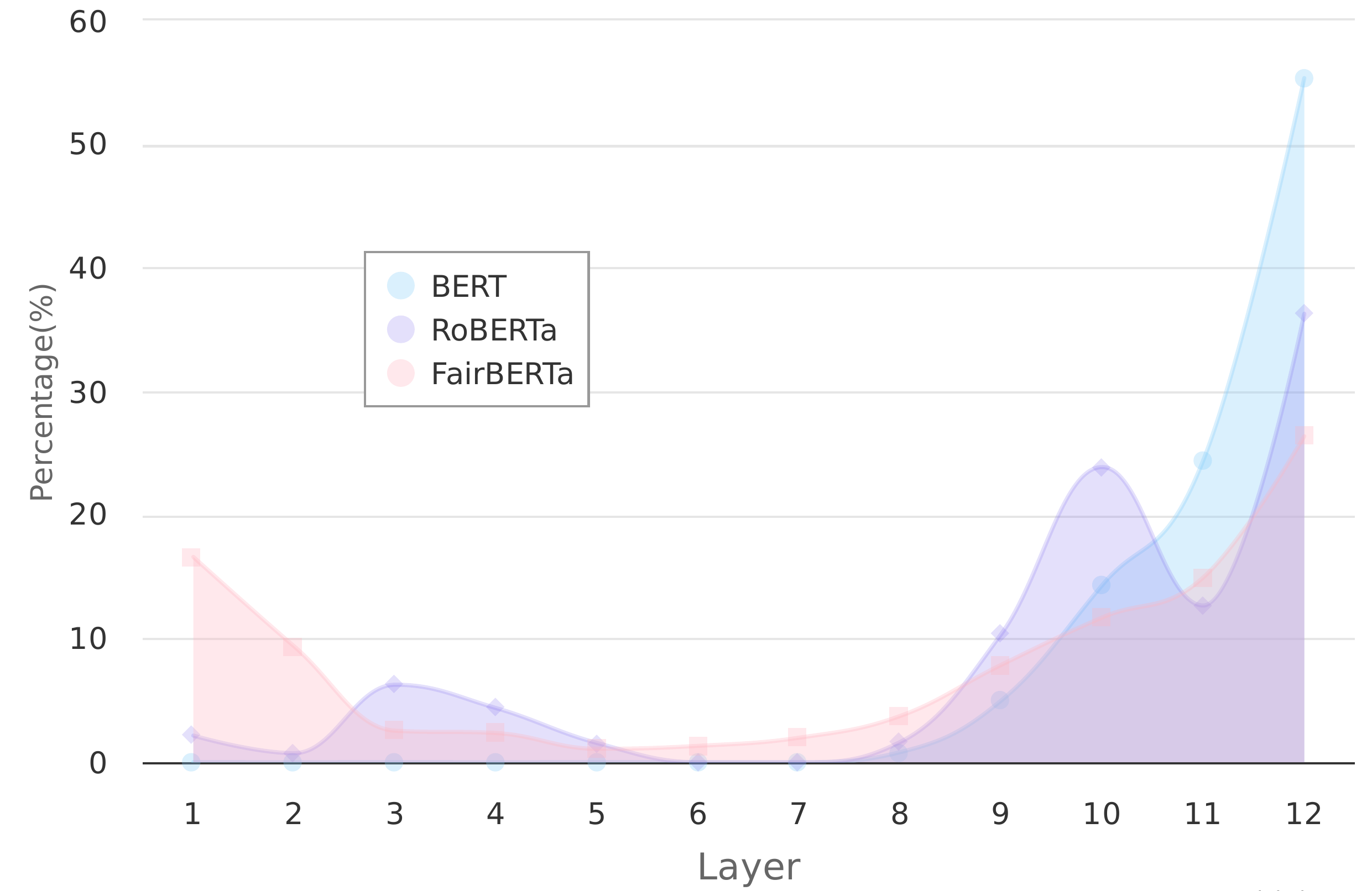}
\caption{\label{kn_distribution_cmp_fig} 
The distribution comparison of pinpointed social bias neurons in each Transformer layer for BERT, RoBERTa, and FairBERTa.
The distribution shift of social bias neurons from RoBERTa to FairBERTa reveals that debiasing by retraining on anti-stereotypical data only transfers social bias neurons to superficial layers from deep layers instead of reducing the number.
}
\end{minipage}
\end{figure}

Table~\ref{debias_res} presents the comparisons with other debiasing methods. 
We observe that Union\_IG, while achieving better debiasing performance ($e.g.$, $53.82$ stereotype score for RoBERTa-Base), severely impairs the language model's capability ($91.70$ $\rightarrow$ $30.61$ of LMS). This is because Union\_IG indiscriminately suppresses all the neurons relating to different demographics, which inevitably damages other useful knowledge learned by the model. 
In contrast, our method BNS maximizes the retention of useful knowledge and only accurately locates neurons that cause distribution gaps for different social groups, achieving a significantly better ICAT score of $84.79$. These distribution gaps are the essential reasons why model outputs contain social biases. 

When compared to other methods such as AutoDebias and FairBERTa, our BNS also performs significantly better. It's worth noting that FairBERTa requires model retraining, whereas our BNS is training-free, only demanding minimal computational resources and being highly efficient.

\vspace{-3mm}

\section{Analysis and Insights}

\subsection{Distribution Shift Of FairBERTa}
Figure~\ref{kn_distribution_cmp_fig} compares the distribution of pinpointed social bias neurons for three pre-trained language models: BERT, RoBERTa, and FairBERTa.
We find that the distribution of most social bias neurons for BERT and RoBERTa are concentrated in the deepest three layers ($10$th, $11$th, $12$th).
By comparing the distribution of social bias neurons for RoBERTa and FairBERTa, we observe that although there has been little change in quantity after debiasing (from $10.35$ to $10.06$ on average), there have been noteworthy alterations in the distribution of these neurons.
Before debiasing, social bias neurons were predominantly concentrated in the deepest three layers ($10$th, $11$th, $12$th), with the highest concentration found in the deepest layer ($12$th), accounting for approximately $55.7\%$ (BERT) and $36.4\%$ (RoBERTa) of the total quantity. 
After debiasing, we notice that a substantial quantity (almost $1/3$) of social bias neurons migrated from the deepest three layers ($10$th, $11$th, $12$th) to the shallowest three layers ($1$st, $2$nd, $3$rd). 
We observe a notable increase in the number of neurons in the two shallowest layers ($1$st, $2$nd), rising from $2.9\%$ (RoBERTa) to $26.0\%$ (FairBERTa). Meanwhile, the number of neurons in the deepest layer ($12$th) decreased to $26.4\%$.
Based on this phenomenon, we speculate that social bias neurons have a considerable adverse impact on the model outputs due to their proximity to the final output layer before debiasing.
By pre-training on the anti-stereotypical data,  FairBERTa transfers the social bias neurons from the deepest three layers to the shallowest three layers to mitigate social biases.
We analyze that the significant reduction of social bias neurons near the final output layer alleviates their impacts on model outputs, which could be the secret lying behind the effectiveness of the debiasing method in FairBERTa.

\begin{table}[t]
\footnotesize
\centering
\resizebox{1\linewidth}{!}{
\begin{tabular}{l c c c c c c c c c c c c c c c}
\toprule
\multirow{2}{*}{\textbf{Model}}
& \multicolumn{3}{c} {\textbf{Ethnicity}}
&
& \multicolumn{3}{c} {\textbf{Physical Appearance}}
&
& \multicolumn{3}{c} {\textbf{Politics}}
\\
\cline{2-4}
\cline{6-8}
\cline{10-12}
& \textbf{Avg. BN}
& \textbf{Avg. Intra}
& \textbf{Avg. Inter} 
&
& \textbf{Avg. BN}
& \textbf{Avg. Intra}
& \textbf{Avg. Inter} 
&
& \textbf{Avg. BN}
& \textbf{Avg. Intra}
& \textbf{Avg. Inter} 
\\ 
\midrule
BERT 
& $14.57$ & $10.97$ & $0.34$ & & $2.91$ & $2.13$ & $0.04$ & & $3.49$ & $2.31$ & $0.01$ \\
RoBERTa 
& $14.05$ & $8.01$ & $0.38$ & & $3.17$ & $1.13$ & $0.02$ & & $4.75$ & $3.06$ & $0.00$ \\
FairBERTa 
& $13.92$ & $8.09$ & $0.28$ & & $3.04$ & $1.27$ & $0.04$ & & $5.13$ & $3.79$ & $0.01$ \\
\bottomrule
\end{tabular}
}
\vspace{-2mm}
\caption{
Statistics of social bias neurons.
``Avg. BN'' means the average number of Bias Neurons pinpointed by our IG$^2$.
``Avg. Intra'' means the average number of neurons in the intersection within the same JUT (only prompt templates are different).
``Avg. Inter'' means the average number of neurons in the intersection for different JUTs (either different UTs or different Judgmental Modifiers).
}
\label{tab:neuron_stt}
\vspace{-3mm}
\end{table}

\subsection{Statistics of Social Bias Neurons}
\begin{wrapfigure}{r}{0.55\textwidth}
\raggedleft
\lstset{
xleftmargin=0.01\linewidth
}
\centering
\includegraphics[width=1.0\linewidth]{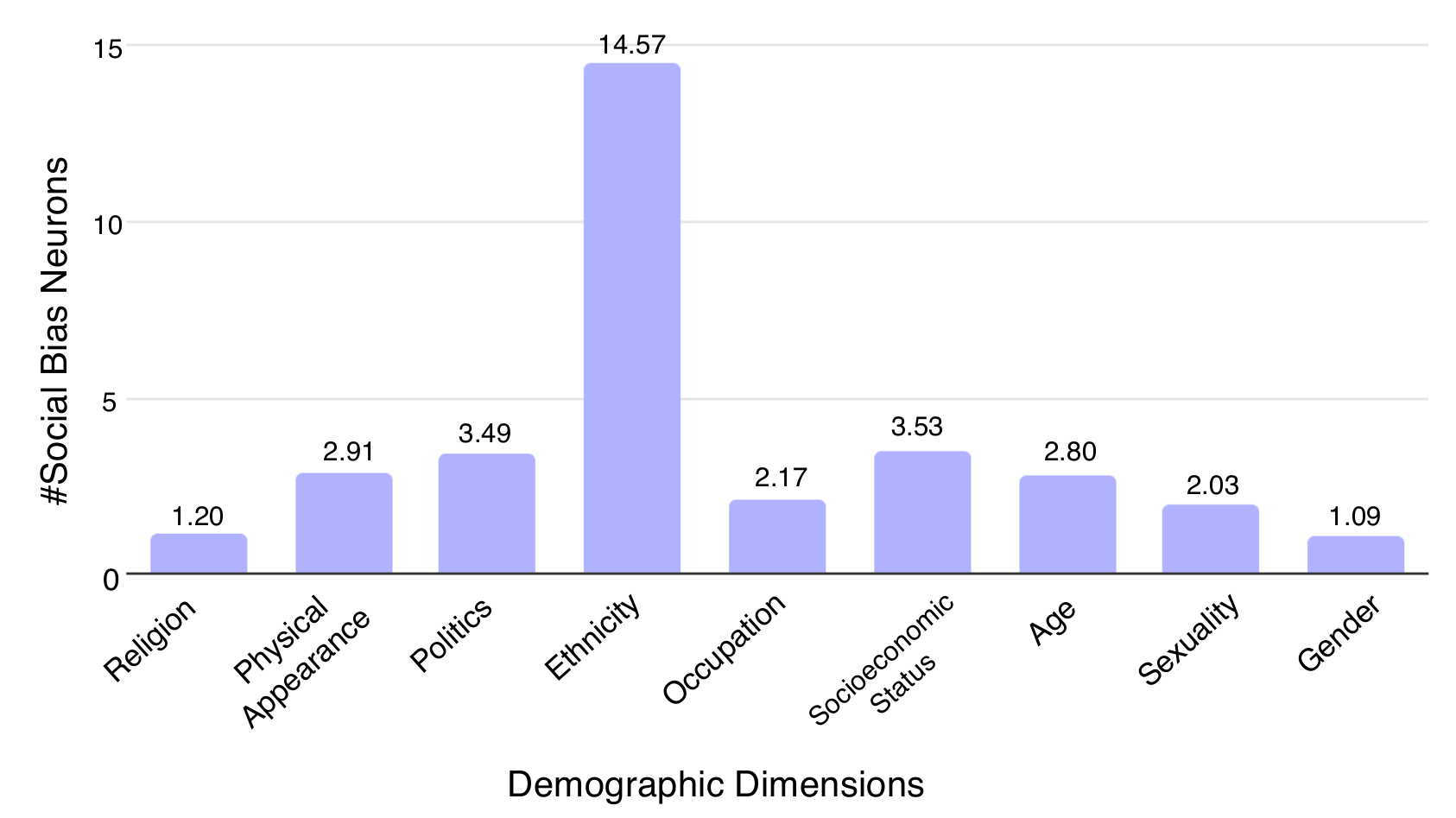}
\caption{\label{neuron_num} 
The average number of social bias neurons pinpointed in BERT for different demographic dimensions. Best viewed on screen.
}
\end{wrapfigure}

Figure~\ref{neuron_num} presents the average number of social bias neurons pinpointed by our IG$^2$ for different demographic dimensions studied in this work.
As shown, the average number varies for different demographic dimensions, with that for ``Ethnicity'' being the most ($14.57$) and that for ``Gender'' being the fewest ($1.09$).
Except for the ``Ethnicity'' dimension, we on average only need to suppress the activation of less than $4$ neurons to effectively mitigate social biases for other dimensions.

Table~\ref{tab:neuron_stt} shows the statistics of social bias neurons.
As we can see, for each demographic dimension, the same JUT shares most of the pinpointed social bias neurons, while different JUTs share almost no social bias neurons.
The large value of Avg. Intra indicates that for the same JUT (confirmed Judgemental Modifier and Unfair Target), pinpointed social bias neurons remain almost the same for different prompt templates.
This further verifies the accuracy of the social bias neurons pinpointed by our IG$^2$, because this suggests that pinpointed neurons are not related to the sentence structure or grammar of the prompt template, but only to the social bias in the studied social group.
The small value of Avg. Inter indicates that social bias neurons pinpointed by our IG$^2$ are quite different for different JUTs, even if prompt templates are the same for them.

\section{Related Work}
Since various AI applications permeate every aspect of our lives~\citep{zan2024codes,10094616,zan2023privatelibraryoriented,chen2022d3etr,liu-etal-2022-mpii,zan2023programming,liu2021enhance}, research on AI Ethics~\citep{liu2023code,liu2023sql,NinarehMehrabi2019ASO} has attracted more and more attention.
In this work, we mainly explore one important aspect of AI Ethics: AI Fairness, which has been studied from different perspectives~\citep{equal_opportunity,john2022reality,stereoset,nangia-etal-2020-crows}.
\citep{HaochenLiu2023TowardAG} proposed to study the existence of annotator group bias in various real-world crowdsourcing datasets. \citep{li-etal-2022-herb} measured hierarchical regional bias in pre-trained language models.
Some works tried to detect and mitigate social biases in word embeddings~\citep{woman_homemaker,kaneko2022gender} and hidden representations~\citep{chowdhury2022learning}, while others explored quantifying social biases in downstream tasks.
Many works have explored the fairness problem in text classification tasks~\citep{dixon2018measuring,liu2021authors,dinan2020multi}. 
Some works also explore the fairness problem in generation tasks, such as machine translation~\citep{stanovsky-etal-2019-evaluating}, story generation~\citep{lucy-bamman-2021-gender}, and question answering~\citep{parrish-etal-2022-bbq}.
However, no work has focused on the interpretability of the fairness research.
In this paper, we close this gap by proposing an interpretable technique specific to the study of social bias along multiple dimensions.

\section{Limitations and Conclusion}
In this paper, we propose a novel interpretable technique, Integrated Gap Gradients (IG$^2$), to precisely identify social bias neurons in pre-trained language models.
We also develop a new dataset to facilitate the interpretability study of social bias.
Derived from our interpretable technique, {\sc Bias Neuron Suppression (BNS)} is further proposed to mitigate social bias.
Extensive experiments have verified the effectiveness of our IG$^2$ and BNS.
In addition, facilitated by our interpretable method, we analyze the distribution shift of social bias neurons after debiasing and obtain useful insights that bring inspiration to future fairness research.

\noindent \textbf{Limitations.} While our study provides valuable insights, we recognize there exist limitations. For example, our proposed BNS method directly sets the activation values of selected social bias neurons to zero. Although this is effective, designing a more refined suppression method might yield even better results. These present opportunities for future research.

\bibliography{iclr2024_conference}
\bibliographystyle{iclr2024_conference}

\end{document}

%% file: math_commands.tex
\usepackage{amsmath,amsfonts,bm}

\def\eqref#1{equation~\ref{#1}}

\def\1{\bm{1}}

\DeclareMathAlphabet{\mathsfit}{\encodingdefault}{\sfdefault}{m}{sl}
\SetMathAlphabet{\mathsfit}{bold}{\encodingdefault}{\sfdefault}{bx}{n}